\documentclass[lettersize,journal]{IEEEtran}
\usepackage{amsmath,amsfonts}
\usepackage{algorithm}
\usepackage{algorithmicx}
\usepackage{algpseudocode}
\usepackage{array}
\usepackage[caption=false,font=normalsize,labelfont=sf,textfont=sf]{subfig}
\usepackage{textcomp}
\usepackage{stfloats}
\usepackage{url}
\usepackage{verbatim}
\usepackage{graphicx}
\usepackage{cite}

\usepackage{booktabs}
\usepackage{multirow}
\usepackage{caption} 
\usepackage{subcaption} 
\usepackage{multirow, multicol}
\usepackage{color}
 \usepackage[colorlinks]{hyperref}

\usepackage{hyperref}

\begin{document}
\title{Colo-ReID: Discriminative Representation Embedding with Meta-learning for Colonoscopic Polyp Re-Identification}
\author{Suncheng Xiang, \IEEEmembership{Member, IEEE}, Chengfeng Zhou, Zhengjie Zhang, Shilun Cai, and Dahong Qian, \IEEEmembership{Senior Member, IEEE}
\thanks{This work was partially supported by the National Natural Science Foundation of China under Grant No.62301315.}
\thanks{Suncheng Xiang, Chengfeng Zhou, Zhengjie Zhang and Dahong Qian are with the School of Biomedical Engineering, Shanghai Jiao Tong University, No 800, Dongchuan Road, Minhang District, Shanghai, China, 200240  (email: xiangsuncheng17@sjtu.edu.cn; joe1chief1993@gmail.com; z876252209@sjtu.edu.cn; dahong.qian@sjtu.edu.cn).}
\thanks{Shilun Cai is with the Endoscopy Center, Zhongshan Hospital of Fudan University, No 180, Fenglin Road, Xuhui District, Shanghai, China, 200032  (email: caishilun1988@qq.com).}}



\markboth{Journal of \LaTeX\ Class Files,~Vol.~14, No.~8, August~2021}%
{Shell \MakeLowercase{\textit{et al.}}: A Sample Article Using IEEEtran.cls for IEEE Journals}


\maketitle

\begin{abstract}
Colonoscopic Polyp Re-Identification aims to match the same polyp from a large gallery with images from different views taken using different cameras and plays an important role in the prevention and treatment of colorectal cancer. However, traditional methods for object ReID directly adopting CNN models trained on the ImageNet dataset usually produce unsatisfactory retrieval performance on colonoscopic datasets due to the large domain gap. Additionally, these methods neglect to explore the potential of self-discrepancy among intra-class or inter-class relations in the colonoscopic polyp dataset, which remains an open research problem in the medical community. To solve this dilemma, we propose a simple but effective training method named \textbf{Colo-ReID}, which can help our model learn more general and discriminative knowledge based on the meta-learning strategy in scenarios with fewer samples. Based on this, a dynamic \textbf{M}eta-\textbf{L}earning \textbf{R}egulation mechanism called \textbf{MLR} is introduced to further boost the performance of polyp re-identification. Our experimental results show that Colo-ReID consistently outperforms second-best method in terms of mAP performance by \textbf{+2.3\%} on polyp re-identification task. Our source code is also publicly available at \url{https://github.com/JeremyXSC/Colo-ReID}.
\end{abstract}


\section{Introduction}
\label{sec1}
\begin{figure}[!t]
\centering{\includegraphics[width=\linewidth]{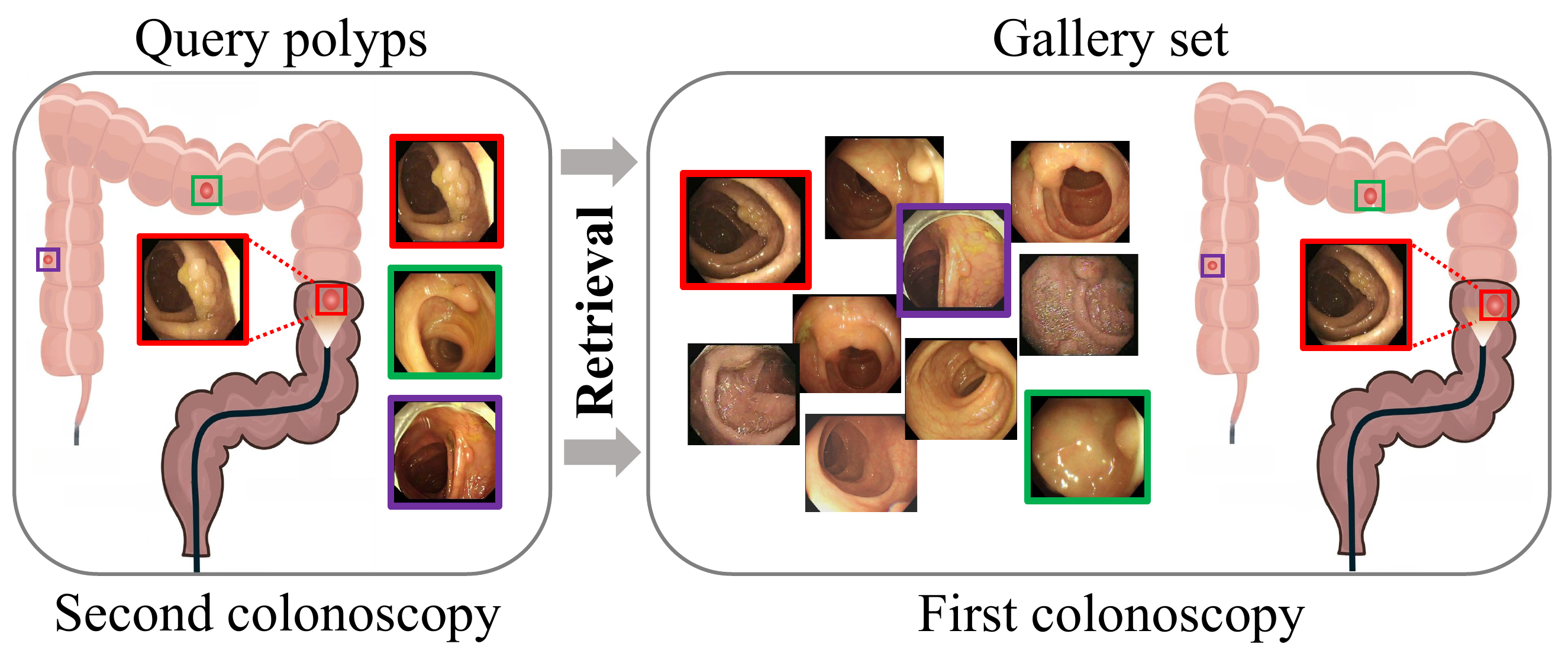}}
\caption{Illustration of polyp re-identification task. Given the query video clip of colonoscopic polyp, colonoscopic video retrieval aims to accurately locate or find the similar clip which semantically corresponds to the given query.}
\label{fig1_pre}
\end{figure}

\IEEEPARstart{C}{olonoscopic} polyp re-identification (Polyp ReID) aims to match a specific polyp in a large gallery with different cameras and locations, the task of which has been studied intensively due to its practical significance in the prevention and treatment of colorectal cancer (CRC) in computer-aided diagnostics. 
As the gold standard for CRC screening, colonoscopies can significantly reduce the risk of death from CRC by early detection of tumors and removal of precancerous lesions~\cite{xu2022deep}. In essence, most large-size polyps cannot be removed during the first procedure. Consequently, missed diagnosis of adenoma may lead to tumor progression, thereby delaying treatment. Inspired by this, physicians are alerted to pay attention to the possible presence of previously identified polyps in the nearby area when the polyps are successfully re-identified, which gives rise to the birth of Polyp ReID task, as illustrated in Fig.~\ref{fig1_pre}. 

With the development of deep convolution neural networks and the availability of video re-identification datasets, video retrieval  methods have achieved remarkable performance in a supervised or unsupervised manner~\cite{xiang2023deep}, where a model is trained and tested on different splits of the same dataset.
However, in practice, manually labelling a diverse set of pairwise polyp area data is time-consuming and labor-intensive when directly deploying the polyp ReID system to new hospital settings~\cite{chen2023colo}. During the intensive annotation process, medical doctors need to associate a
colonoscopic polyp across different cameras, which is a difficult and laborious process as colonoscopic polyp data might exhibit very different appearances when captured by different cameras. 



\begin{figure}[!t]
\centering{\includegraphics[width=\linewidth]{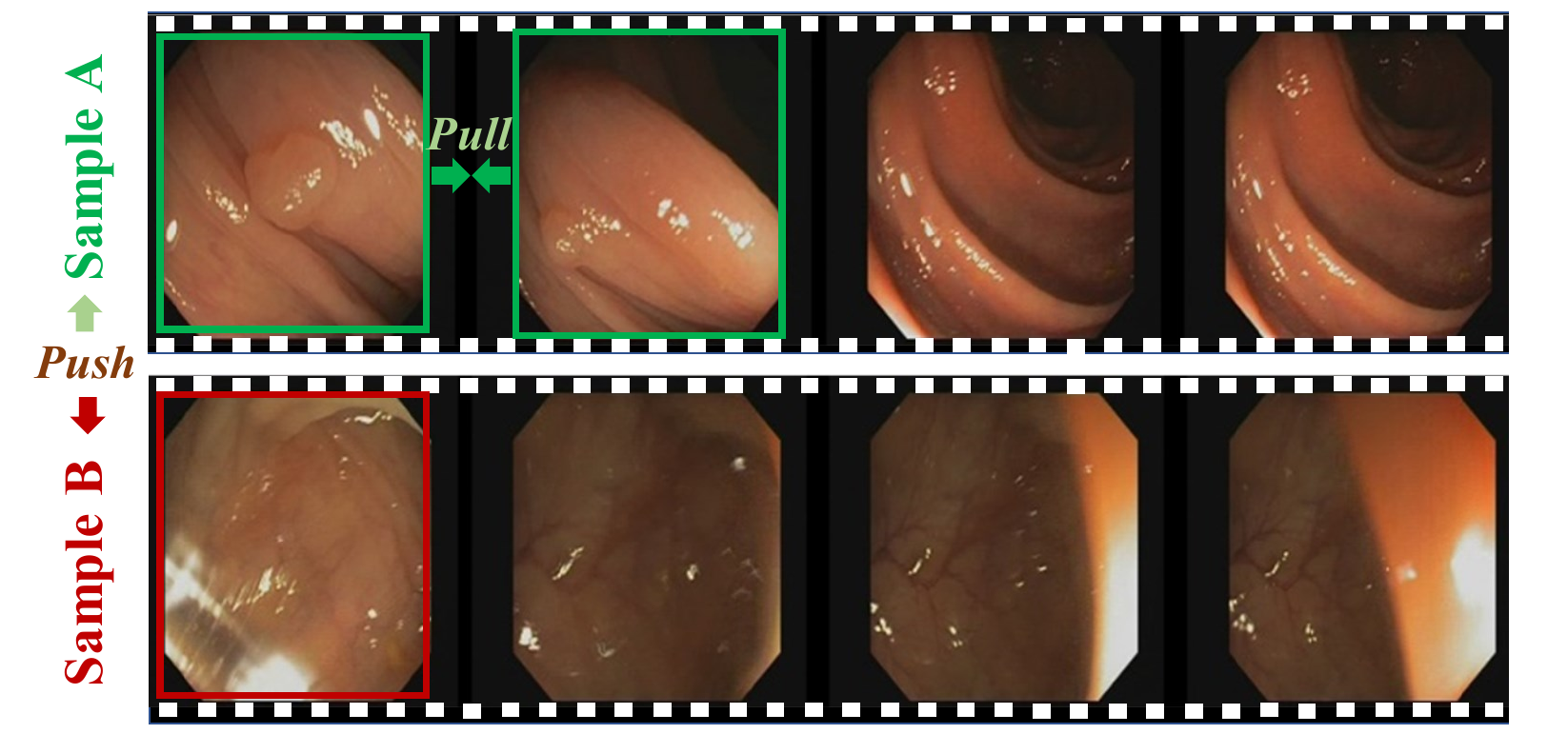}}
\caption{Description of the self-discrepancy of intra-class or inter-class relations for polyp dataset, which is caused the experience variations of the surgeon doctors, the variation in the shooting environment and the difference in patient condition, etc.}
\label{fig1}
\end{figure}

To facilitate research in this area, several works~\cite{azagra2022endomapper,chen2023colo} have proposed to collect manually-constructed datasets of complete colonoscopy sequences for colonoscopic polyp ReID. For example,
Azagra et al.~\cite{azagra2022endomapper} introduced the first collection of complete endoscopy sequences named Endomapper for endoscopic visual simultaneous localization and mapping. Recently,
Chen et al.~\cite{chen2023colo} constructed a fine-grained colonoscopic dataset named Colo-Pair for colonoscopic video retrieval.
However, all aforementioned datasets are still on a limited scale due to the expensive time costs and intensive human labors from medical doctors and as such, fail to meet the standard requirements of large-scale deep training of neural networks. Considering this, a new learning paradigm is urgently needed for few-shot scenarios.
Another possible challenge is that, as illustrated in Fig.~\ref{fig1}, there are still some self-discrepancies of intra-class or inter-class relations among colon polyp datasets for the colonoscopic polyp ReID task due to the variation in terms of camera environment, viewpoint, illumination, and so forth. Previous methods mainly focus on improving model robustness and discriminability by model optimization, while neglecting to explore the self-discrepancy of intra-class or inter-class relations of colonoscopic polyp datasets, which hinders further performance improvement for the polyp re-identification task.


In order to alleviate the above-mentioned problems in terms of data scale and learning paradigm, we adopt the concept of ``learning how to learn" in meta-learning to simulate the training and testing procedure in the process of model optimization for the polyp re-identification task. We innovatively propose a meta-learning based polyp ReID model for scenarios with few samples. To be more specific, we creatively construct a Colo-ReID model for more robust visual-semantic embedding for polyp ReID task. On the basis of it, a dynamic \textbf{M}eta-\textbf{L}earning \textbf{R}egulation mechanism (\textbf{``MLR" for short}) is introduced to dynamicly optimize the parameters of the model, so that new proposed polyp ReID model can adapt to new tasks in scenarios with small amounts of training data.


Our model is trained on the collected colonoscopic polyp images from the source domain, and aims to generalize to any unseen scenarios for effective polyp ReID without utilizing any model fine-tuning strategies.
During the training process, we adopt triplet loss as the optimization function,
which can effectively exploit the self-discrepancy of intra-class or inter-class relations of colonoscopic polyp datasets in few-shot scenarios to significantly boost model performance on the polyp ReID task.
During the testing process, our model adopts meta-learning normalization to generate more diverse meta-measurement features, which allows it to be more robust in the general scenarios.
To the best of our knowledge, this is the first research effort to exploit the potential of self-discrepancy among intra-class relations in colonoscopic polyp datasets on the polyp re-identification task.
Compared with traditional polyp re-identification approaches, our Colo-ReID method is different from two perspectives:
1) Firstly, previous methods require large-scale datasets for model training to obtain competitive retrieval performance while our Colo-ReID model works on a wide range of retrieval tasks in few-shot scenarios, which allows our model to be more flexible and adaptable in real-world scenarios;
2) Secondly, our proposed Colo-ReID method is designed to take advantage of the self-discrepancy of intra-class or inter-class relations among colonoscopic polyp datasets. In contrast, previous methods neither explore these characteristics in real-world scenarios, nor benefit vastly from sub-optimal or meta-learning based results, which limits their practicality in real-world scenarios.



As a consequence, the major contributions of our work can be summarized into three-fold:
\begin{itemize}

 \item[$\bullet$] We propose a simple but effective training method named Colo-ReID, which helps our model to learn general knowledge based on the meta-learning strategy.
 
 \item[$\bullet$] Based on it, a dynamic \textbf{M}eta-\textbf{L}earning \textbf{R}egulation mechanism called \textbf{MLR} is introduced to further boost the performance on the colonoscopic polyp ReID task.

 \item[$\bullet$] Comprehensive experiments show that our Colo-ReID method matches or exceeds the performance of existing methods with a clear margin, which reveals the applicability of colonoscopic polyp ReID task with new insights.
\end{itemize}

In the rest of the paper, we first review some related works of colonoscopic polyp re-identification methods and previous semantic-based methods in Section \ref{sec2}. Then in Section \ref{sec3}, we give more details about the learning procedure of the proposed Colo-ReID method. Extensive evaluations compared with state-of-the-art methods and comprehensive analyses of the proposed approach are elaborated in Section \ref{sec4}. Conclusion and future works are given in Section \ref{sec5}.

\section{Related Works}
\label{sec2}
In this section, we give a brief review of the related works using common re-identification methods. The core idea of these existing methods is to learn a discriminative and robust model for downstream object re-identification. These methods can be roughly divided into hand-crafted based approaches and deep learning based approaches.

\subsection{Hand-crafted based Approaches}
Historically, the success of machine learning has been driven by the advancement of hand-engineered features.
Traditional studies~\cite{zhao2014learning,wang2007shape} related to hand-crafted systems for image retrieval aim to design or learn robust features. For example, Zhao et al.~\cite{zhao2014learning} explored different discriminative abilities of local patches, and proposed an approach of learning mid-level filters from patch clusters for person re-identification. Wang et al.~\cite{wang2007shape} proposed an appearance model that captures the spatial relationships among appearance labels and prove that it can be used to dramatically quicken occurrence computation. In light of recent advances in image search,
Zheng et al.~\cite{zheng2015person} designed an unsupervised Bag-of-Words (BoW) representation to bridge the gap
between image search and person re-identification.
Unfortunately, these models fail to show competitive performance since these hand-crafted based approaches have some non-negligible limitations. On the one hand, they cannot jointly optimize for feature extraction and metric learning since these two inter-connected components are independently designed and cascaded. On the other hand, these early works are mostly based on heuristic design and thus could not learn optimal discriminative features on current large-scale datasets.

\subsection{Deep Learning based Approaches}
Recently, there has been significant research interest in the design of deep learning based approaches for video retrieval. For example, 
Intrator et al.~\cite{intrator2023self} proposed a multi-view self-supervised learning method for learning informative representations of a sequence of video frames, which can be used to group disjoint tracklets generated by a spatio-temporal tracking algorithm based on their appearance. 
Kordopatis et al.~\cite{kordopatis2022dns} proposed a video retrieval framework based on knowledge distillation that addresses the problem of performance-efficiency trade-off.  Yu et al.~\cite{yu2022end} present a plug-in module named instance tracking head named ITH for synchronous polyp detection and tracking, which can be assembled on an object detector for real-time detection and tracking of polyp instances in colonoscopy videos.
Chen et al.~\cite{chen2023colo} also proposed a self-supervised contrastive representation learning scheme named Colo-SCRL to learn spatial representations from colonoscopic video datasets, which demonstrates the priority of visual representation learning.

\begin{figure*}[!t]
\centering{\includegraphics[width=0.9\linewidth]{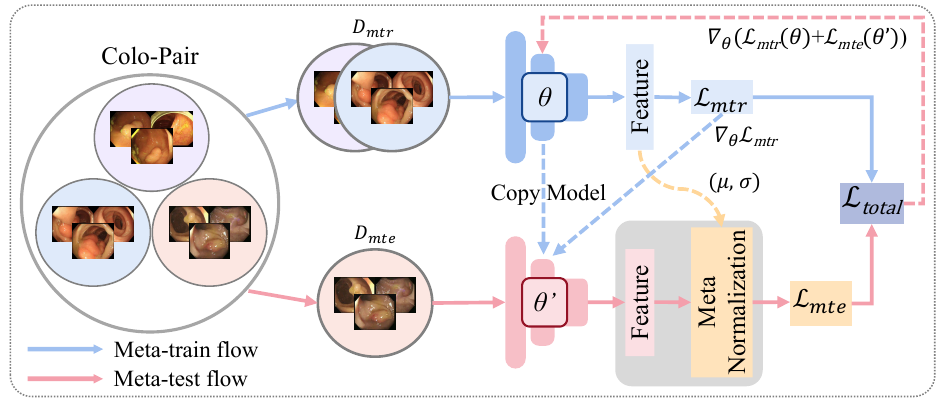}}
\caption{Overview of the proposed Colo-ReID framework, which can randomly divide the sampled data into meta training set $\mathcal{D}_{mtr}$ and meta testing set $\mathcal{D}_{mte}$ respectively, and then enable our polyp ReID model to learn more robust and discriminative
knowledge with the limited colonoscopy polyp matching data, which can significantly enhance the discriminative ability of our model.}
\label{fig2}
\end{figure*}

As for the meta-learning methods, Yu et al.~\cite{yu2022meta} propose Meta-ADD based on Meta learning, which can learn to classify concept drift by offline pre-training a model on data stream with known drifts,
then online fine-tuning model to improve detection accuracy. Khodak et al.~\cite{khodak2019adaptive} introduce ARUBA framework for designing and understanding practical meta-learning methods that integrates sophisticated formalizations of task-similarity with the extensive literature on online convex optimization and sequential prediction algorithms. Besides,
Wang et al.~\cite{wang2022meta} propose a meta-learning based hyperspectral target detection using Siamese network to enhance the generalization ability of the transferred model and its adaptation to new tasks.
Typically, these approaches first rank videos based on a coarse-grained manner, in order to filter videos with similarity lower than a predefined threshold. However, learning visual representation is by no means a trivial task. In addition, in these approaches, both coarse- and fine-grained components are typically built based on hand-crafted features with traditional aggregations and heuristic/non-learnable approaches for similarity calculation which results in sub-optimal performance.
To this end, these solutions are incapable of capturing a large variety of temporal similarity patterns due to their rigid aggregation approach, which hinders the further performance improvement for colonoscopic polyp retrieval tasks.

To address the challenges mentioned above, we propose a novel meta-learning based model named \textbf{Colo-ReID} to improve performance on the polyp re-identification task. Based on it, a \textbf{M}eta-\textbf{L}earning \textbf{R}egulation mechanism named \textbf{MLR} is introduced into the meta-training and meta-testing phase, which can significantly improve the performance and robustness of the Colo-ReID method for the colonoscopic polyp ReID task. Specifically, the model adopts the strategy of second-order gradient updating for meta-train and meta-test phases to polyp re-identification task, then significantly reduces the interference time of internal differences in training data. At the same time, a meta-normalization mechanism is used in the meta-test phase to further diversify the meta-test features and alleviate limitations of the small sample problem on model performance, which helps to significantly capture common features in the colonoscopy video and improve the stability of the model. To the best of our knowledge, this is the first attempt to employ the meta-learning paradigm for the colonoscopic polyp ReID task. We hope this paper will help popularize the use of meta-learning based approaches in medical image analysis yielding improvements in representation learning efficiency across the medical field.


\section{Methodology}
\label{sec3}

\subsection{Preliminary}
We begin with a formal description of the colonoscopic polyp re-identification (Polyp ReID) problem. Assuming that we are given a source domain $\mathcal{D}$, which contains its own image-label pairs $\mathcal{D}=\left\{\left(\boldsymbol{x}_i, y_i\right)\right\}_{i=1}^{N}$ of colonoscopic videos, where  $N$ is the number of images in the source domain $ \mathcal{D}$. Each sample $\boldsymbol{x}_i \in \mathcal{X}$ is associated with an identity label $y_i \in \mathcal{Y}=\left\{1,2, \ldots, M\right\}$, where $M$ is the number of identities in the source domain $\mathcal{D}$. 
The goal of this paper is to leverage labeled source training polyp samples to learn the discriminative embeddings of the target testing set on polyp ReID task.

\subsection{Our Proposed Colo-ReID Method}

In this section, we will give further details about the proposed Colo-ReID method on the basis of the meta-learning strategy. To further explore the self-discrepancy of intra-class or inter-class relations in colonoscopic polyp datasets and alleviate the few-shot learning problem, we adopt the meta-learning strategy (\textit{``learning to learn"}) to simulate the training and testing process on the polyp ReID task during model optimization. Then, we introduce the two-stage process of computing the meta-learning loss (\textbf{meta-training} and \textbf{meta-testing}) as well as the meta-learning training strategy based on quadratic gradient updating. Finally, we will give a detailed explanation about our training optimization process.


According to the preliminary analysis, there exists obvious self-discrepancy of intra-class or inter-class relations in the colonoscopic polyp datasets, which are mainly caused by variations in camera environment, viewpoint, illumination and the difference between different patients, amongst other factors. To solve this dilemma, we introduce a meta-learning based strategy into the training process to better capture the common features in colonoscopy videos, which can significantly improve the robustness of our model.

In the area of medical retrieval task, recent methods mainly focus on obtaining more discriminative feature embeddings on the basis of the deep neural network. Intuitively, different image retrieval tasks have certain commonalities in their common feature space. Based on this assumption, our meta-learning strategy can sample data from different domains to form different tasks.
Through a small number of gradient descent steps, the Colo-ReID model can obtain initialization parameters that can quickly adapt to new tasks.
Inspired from the merits of the meta-learning techniques, we propose the Colo-ReID model to improve the performance of colon polyp re-identification without the collection of large-scale visual-text data catered towards the polyp re-identification task.
As illustrated in Fig.~\ref{fig2},
our proposed Colo-ReID method selects data from different domains, and randomly divides the sampled data into a meta-training set $\mathcal{D}_{mtr}$ and a meta-testing set $\mathcal{D}_{mte}$.
The common characteristics of the polyp re-identification task are fully mined from the limited polyp-matching data to improve the discrimination and matching ability of the model, which can effectively help our model learn general knowledge with fewer samples.
To be more specific, we generate the gradient feedback for updating the initial parameters $\theta$, which can be described as:
\begin{equation}
\begin{aligned}
 \min _\theta \sum_{T_i \sim p(T)}  &\mathcal{L}\left(\theta-\alpha \nabla_\theta \mathcal{L}\left(\theta, D_{T_i}^{m t r}\right), D_{T_i}^{m t e}\right)  \\ &=\min _\theta \sum_{T_i \sim p(T)} \mathcal{L}\left(\theta, D_{T_i}^{m t e}\right) \end{aligned}
\label{eq1}
\end{equation}
where $T$ indicates the target domain, while $p(T)$ denotes the probability of subsets sampled from target source; and $L$ represents the full objective loss for our optimization, which is already defined in Eq.~\ref{eq9}. Besides, $D_{T_i}^{m t r}$ and $D_{T_i}^{m t e}$ represent the meta-training set and meta-testing set from $i$-th target domain respectively.

In continuation, our model is trained on the meta-training set $\mathcal{D}_{mtr}$, and then tested on the meta-testing set $\mathcal{D}_{mte}$ to validate the effectiveness of our training strategy.
The parameter-updating process is illustrated in Fig.~\ref{fig3}. To be more specific, we randomly divide the sample set $\mathcal{D}$ into meta-training set $\mathcal{D}_{mtr}$ and the meta-testing set $\mathcal{D}_{mte}$, then
we compute the meta-training loss $\mathcal{L}_{mtr}$ on the meta-training samples in the meta-training phase, which can be expressed as follows:

\begin{figure}[!t]
\centering{\includegraphics[width=1.0\linewidth]{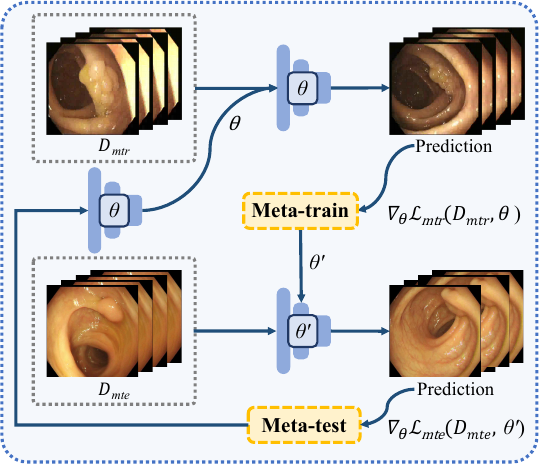}}
\caption{Schematic of the parameter update process using meta-training and meta-testing. Specifically, we randomly divide the sample set $\mathcal{D}$ into meta-training set $\mathcal{D}_{mtr}$ and the meta-testing set $\mathcal{D}_{mte}$. Additionally,
we compute the meta-training loss $\mathcal{L}_{mtr}$ on the meta-training samples in the meta-training phase, and we also 
compute the meta-testing loss $\mathcal{L}_{mte}$ on the meta-testing samples in the meta-testing phase.}
\label{fig3}
\end{figure}


\begin{figure}[!t]
\centering{\includegraphics[width=\linewidth]{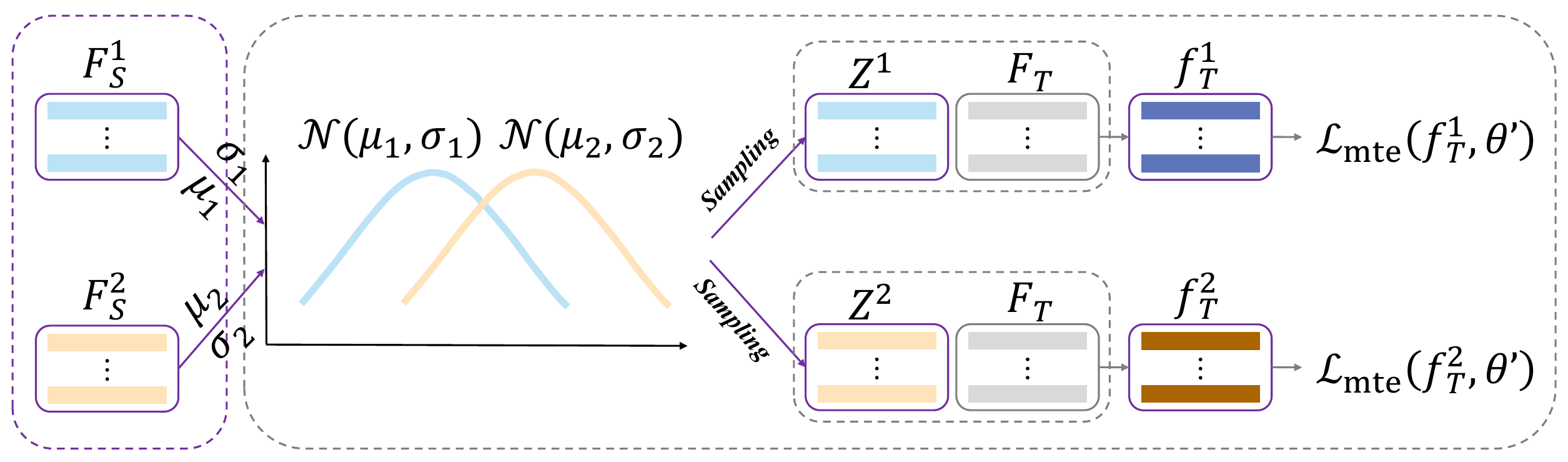}}
\caption{Illustration of our Meta-Learning Regulation mechanism. Specifically, MLR generates more diverse
meta-test features at the feature level, then replaces the last batch normalization layer of the model with a meta-learning regularization layer. Zoom in for the best view.}
\label{fig4}
\end{figure}

\begin{equation}
\mathcal{L}_{m t r}=\mathcal{L}\left(D_{m t r}, \theta\right)
\label{eq2}
\end{equation}
where $\mathcal{L}$ represents the triplet loss, and $\theta$ indicates the parameters of the model.
Then we apply the  meta-train loss $\mathcal{L}_{m t r}$ to perform the first optimization on our model.
\begin{equation}
\theta^{\prime} \leftarrow \operatorname{Adam}\left(\nabla_\theta \mathcal{L}_{m t r}(\theta), \theta, \alpha\right)
\label{eq3}
\end{equation}
where $\theta^{'}$ represents the model parameters optimized by $\mathcal{L}_{m t r}$; and $\alpha$ indicates the learning rate during the meta-training phase.
During the meta-testing phase, we calculate the meta-test loss $\mathcal{L}_{m t e}$ on the meta-test samples using the optimized model, which can be expressed as follows:
\begin{equation}
\mathcal{L}_{m t e}=\mathcal{L}\left(D_{m t e}, \theta^{\prime}\right)
\label{eq4}
\end{equation}

%
%

To summarize, the total optimization goal in this work can be expressed as:
\begin{equation}
\mathcal{L}_{opt} = \underset{\theta}{\operatorname{argmin}} \left( \mathcal{L}_{m t r}(\theta)+\mathcal{L}_{m t e}\left(\theta^{\prime}\right) \right)
\label{eq7}
\end{equation}

Many previous works~\cite{xiang2022learning} have found that performing training with triplet loss has great potential to learn a robust and discriminative model in object re-identification tasks. The triplet loss optimizes the embedding space such that data points with the same identity are closer to each other than those with different identities, which has a significant impact on model performance. Motivated by this, we adopt hard triplet loss to mine the relationship of training samples during training, which can minimize the distance among positive pairs and maximize the distance between negative pairs. Our loss function is defined as:

\begin{equation}
\mathcal{L}_{triplet}=\left(d_{a, p}-d_{a, n}+m\right)_{+}
\label{eq8}
\end{equation}
where $d_{a, p}$, $d_{a, n}$ denote the feature distances of positive pair and negative pairs, respectively, $m$ represents the margin of our triplet loss and $(z)_{+}$ denotes \textit{max(z, 0)}.
To this end, we propose to jointly learn robust visual-semantic embedding using classification loss and triplet loss in a training batch, which can be expressed as:

\begin{equation}
\mathcal{L}= \mathcal{L}_{ID} + \mathcal{L}_{Triplet}
\label{eq9}
\end{equation}
where $\mathcal{L}$, $\mathcal{L}_{ID}$ and $\mathcal{L}_{Triplet}$ represent the toal loss, ID loss and triplet loss, respectively, and the ID loss $\mathcal{L}_{ID}$ is adopted to perform classification on the basis of visual feature.

It is worth mentioning that the meta-test loss $\mathcal{L}_{m t e}$ is only used for updating the $\theta$, and is based on the results of meta-train loss $\mathcal{L}_{m t r}$.
The meta-test loss encourages optimization of the loss on meta-training samples, pushing the model to optimize towards improving accuracy on meta-test samples. Through iterative generalization from the meta-train domain to the meta-test domain, our model can avoid the overfitting problem or domain bias to learn colonoscopy domain-invariant representations with good generalization ability from small-sample data. In other words, by using this meta-learning paradigm, our Colo-ReID model can improve its generalization ability by extracting common knowledge within limited colonoscopy samples, avoiding the overfitting problem to specific domains, mitigating the interference of internal data variations on the Colo-ReID model, and thus enhancing its polyp recognition capability.
Consequently, it contains some discriminative information about the gradient update of the meta-training phase.

\subsection{Meta-Learning Regulation Mechanism}
In our meta-learning strategy, meta-test loss $\mathcal{L}_{m t e}$ is crucial for generalizable feature representation learning. Intuitively, if the meta-test samples are collected from a wider range of colonoscopy distributions, the deep model should be optimized to be more robust and have more stable polyp recognition ability in unseen domains with internal variations. Inspired by this, we employ a dynamic
\textbf{M}eta-\textbf{L}earning \textbf{R}egulation mechanism called \textbf{MLR} to generate more diverse meta-test features at the feature level, which replaces the last batch normalization layer of the model with a meta-learning regularization layer. More details about the whole training process is illustrated in Fig.~\ref{fig4}.
In essence, meta-learning regularization generates more diversified samples for generalizable feature representation learning, then explores the domain-specific information from the meta-training phase to promote representation learning in the meta-testing phase, which can significantly enable the model to learn more robust variations in terms of feature space, as well as mitigating the limitations of the few sample problem for further performance improvement.

\renewcommand{\algorithmicrequire}{ \textbf{Input:}} 
\renewcommand{\algorithmicensure}{ \textbf{Output:}} 
\floatname{algorithm}{\textbf{Algorithm}}
\begin{algorithm}[!t]
\caption{The training procedure of proposed method.}
\small
\label{alg1}
\begin{algorithmic}
 \Require
 Training Dataset $\mathcal{D}$;
Meta-train learning rate $\alpha$;
Meta-test learning rate $\beta$;
Initialized deep model $\theta$;
Maximum training iterations $n$;
 \Ensure 
Optimized polyp ReID model $\theta_{opt}$ 
\label{ code:fram:extract }
\end{algorithmic}
\begin{algorithmic}[1]
\State \textcolor[rgb]{0.11,0.21,0.65}{$\rhd$ \textit{Meta-Learning Process ***}}
\State Initialize: \textit{iter1} = 1;
\label{code1}
\While{$iter1 \leq n $} \do \\
\label{code2}
\State Sample a mini-batch $\mathcal{D}_{b}$ from $\mathcal{D}$;
\label{code3}
\State Random split $\mathcal{D}_{b}$ into $\mathcal{D}_{mtr}$ and $\mathcal{D}_{mte}$;
\label{code4}
\State Copy the original model with $\theta$;
\label{code5}
\State Compute $\mathcal{L}_{m t r}(\theta)$ and $\nabla_\theta \mathcal{L}_{m t r}(\theta)$ with $\mathcal{D}_{m t r}$;
\label{code6}
\State Update $\theta$ to $\theta^{\prime}$ by $\nabla_\theta \mathcal{L}_{m t r}(\theta)$ with the learning rate $\alpha$;
\State Sample features $\mathcal{Z}$ from meta-train distributions;
\label{code7}
\State Generate new meta-test features $f_{T}$ with $\mathcal{D}_{m t e}$ and $\mathcal{Z}$;
\label{code8}
\State Compute $\mathcal{L}_{m t e}(\theta^{\prime})$ and $\nabla_\theta \mathcal{L}_{m t e}(\theta^{\prime})$ with $f_{T}$;
\label{code9}
\State Update $\theta$ by $\nabla_\theta \mathcal{L}_{m t r}(\theta)$ and  $\nabla_\theta \mathcal{L}_{m t e}(\theta)$ with  $\beta$;
\EndWhile

\State \textcolor[rgb]{0.11,0.21,0.65}{$\rhd$ \textit{Meta-Learning Regularization Process ***}}

\State Initialize: \textit{iter2} = 1;
\label{code1}
\While{$iter2 \leq n $} \do \\
\label{code2}
\State Sample a mini-batch $\mathcal{D}_{b}$ from $\mathcal{D}$;
\label{code3}
\State Random split $\mathcal{D}_{b}$ into $\mathcal{D}_{mtr}$ and $\mathcal{D}_{mte}$;
($D_{m t r} \cup D_{m t e}=D, D_{m t r} \cap D_{m t e}=\emptyset$);
\label{code4}
\State Copy the original model with $\theta$;
\State Compute the meta-train loss $\mathcal{L}_{m t r}=\mathcal{L}\left(D_{m t r}, \theta\right)$ in Eq.~\ref{eq2};
\label{code6}
\State Copy the original model and update the copied parameters: $\theta^{\prime} \leftarrow \operatorname{Adam}\left(\nabla_\theta \mathcal{L}_{m t r}(\theta), \theta, \alpha\right)$;
\label{code7}
\State Compute the meta-test loss $\mathcal{L}_{m t e}=\mathcal{L}\left(D_{m t e}, \theta^{\prime}\right)$;
\label{code8}
\State Compute gradient: $g=\nabla_\theta\left(\mathcal{L}_{m t r}(\theta)+\mathcal{L}_{m t e}\left(\theta^{\prime}\right)\right)$;
\label{code9}
\State Update the original model parameters: $\theta \leftarrow \operatorname{Adam}(g, \theta, \beta)$;
\EndWhile
\end{algorithmic}
\end{algorithm}

During the meta-training phase, meta-learning regularization normalizes the meta-training features and saves the batch mean $u_{i}$ and batch variance $\sigma_{i}$ for the $i$-th meta-training domain. In the meta-test phase, meta-learning regularization forms $N_{s}$ Gaussian distributions using the optimized batch mean $\mu_{i}$ and batch variance $\sigma_{i}$.
Each saved mean and variance is calculated with tens of polyp samples, so the generated distribution mainly reflects the high-level domain information rather than the specific polyp information. Based on this, we sample features from these distributions and push these domain-specific features into the meta-test features. Specifically, we sample features from $N_{s}$ meta-training distributions for each mini-batch of meta-test features, \textit{e.g.} for the $i$-th meta-training distribution and the $j$-th meta-test feature, the sampled feature $z_j^i$ can be represented as:

\begin{equation}
z_j^i \sim \mathcal{N}\left(\mu_i, \sigma_i\right)
\label{eq10}
\end{equation}
where $N$ is the Gaussian distribution. Through such sampling, we get $B$ sampled features (where $B$ is also the batch size of meta-test features), which are mixed with the original meta-test features to generate new features $F_T^i$:
\begin{equation}
F_T^i=\lambda * F_T+(1-\lambda) Z^i
\label{eq10}
\end{equation}
where $F_T$ is the original meta-testing feature, $Z^i=\left[z_0^i, z_1^i, \cdots, z_B^i\right]$ represents the $B$ sampled feature from the  $i-$th Gaussian distribution, and $\lambda$ denotes the mixing coefficient sampled from the $\beta$ distribution, $\lambda \sim Beta(1,1)$.

During the inference period, we adopt the trained Colo-ReID model for down-stream polyp retrieval task. To be more specific, given a query image, polyp re-identification (Re-ID) aims to match the polyp-of-interest across multiple non-overlapped cameras distributed in different places, which can collect large-scale gallery samples in the hospital. The sample with the minimum distance will be the target image we are looking for. In this work, we mainly adopted a trained model on the basis of triplet loss $L_{triplet}$ and ID loss $L_{ID}$ for inferencing to get better retrieval performance in this task.

To summarize, this study investigates the polyp ReID problem and then proposes a simple but effective training method named Colo-ReID, which enables our model to learn more domain invariant knowledge through the meta-learning strategy in a low-sample scenario. Additionally, a dynamic \textbf{M}eta-\textbf{L}earning \textbf{R}egulation  mechanism called \textbf{MLR} is introduced to generate more diverse meta-test features at the feature level to boost performance on the polyp ReID task. Specifically, the inner loop of the meta-learning mainly contains meta-training loss $\mathcal{L}_{m t r}$
and meta-testing loss $\mathcal{L}_{m t e}$, while the outer loop of meta-learning includes triplet loss $\mathcal{L}_{triplet}$ and classification loss $\mathcal{L}_{ID}$. All these loss functions play an important role for the meta-learning in polyp re-identification task.
More training details can be available in Algorithm~\ref{alg1}.
To the best of our knowledge, this method provides a new learning paradigm for the medical image retrieval task to the computer vision community and is also the first attempt to employ meta-learning for the polyp re-identification task.

\begin{figure*}[!t]
\centering{\includegraphics[width=0.9\linewidth]{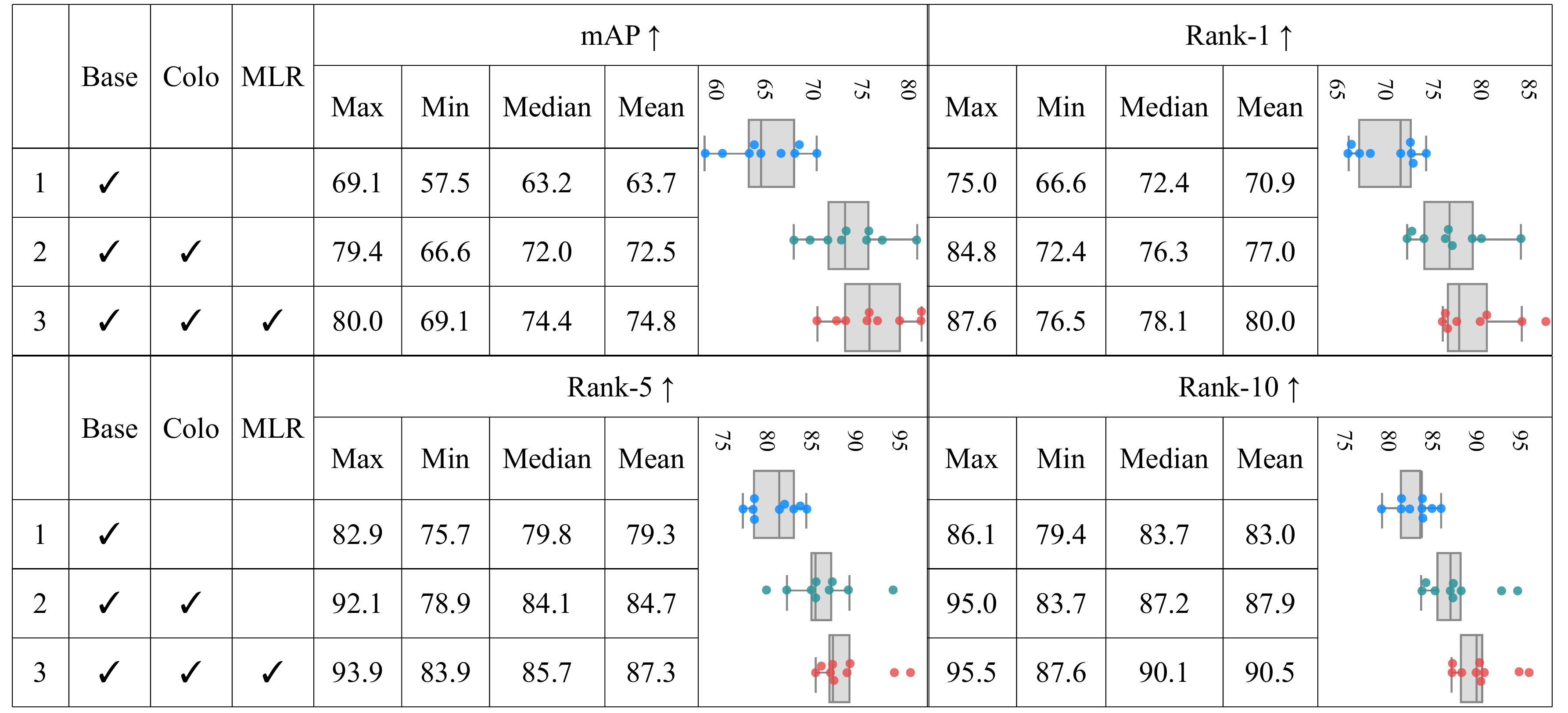}}
\caption{Ablation results on our proposed method. "Base", "Colo" and "MLR" represents the baseline, proposed Colo-ReID method and Meta-Learning Regulation Mechanism, respectively. And  "Mean", "Max", "Median" and "Min" indicate the average, maximum, median and Minimum performance during the 9 repeated experiment. Measured by \%. All p-values $<$ 0.05 (mAP: p$^{12}$ = 6.86 × 10$^{-3}$, p$^{23}$ = 2.82 × 10$^{-5}$, p$^{13}$ = 6.45 × 10$^{-7}$; Rank-1: p$^{12}$ = 1.85 × 10$^{-2}$, p$^{23}$ = 6.55 × 10$^{-5}$, p$^{13}$ = 1.22 × 10$^{-4}$; Rank-5: p$^{12}$ = 3.20 × 10$^{-2}$, p$^{23}$ = 3.46 × 10$^{-4}$, p$^{13}$ = 1.27 × 10$^{-4}$; Rank-10: p$^{12}$ = 2.14 × 10$^{-2}$, p$^{23}$ = 6.67 × 10$^{-4}$, p$^{13}$ = 5.88 × 10$^{-5}$).
}
\label{tab1}
\end{figure*}

\section{Experimental Results}
\label{sec4}
\subsection{Datasets}
\label{sec4.1}

\textbf{Colo-Pair}~\cite{chen2023colo} is the first collection of complete paired colonoscopy sequences, which contains 60 videos from 30 patients, with 62 query video clips and the corresponding polyp clips from the second screening manually annotated as positive retrieval clips,
note that the Colo-Pair dataset is also adopted for four-fold cross validation in this experiment. To be more specific, 2,000 image samples are adopted for testing, while remaining 6,000 image samples are employed for training.
Even though the scale of our Colo-Pair dataset is limited, there are no datasets constructed for this polyp re-identification task so far. In contrast, our newly-built dataset can effectively alleviate this problem. More importantly, our Colo-Pair dataset is still the largest dataset so far in the polyp re-identification scenario. \\
\textbf{Market-1501}~\cite{zheng2015scalable} consists of 32,668 person images of 1,501 identities observed under 6 different camera views. The dataset is split into 12,936 training images of 751 identities and 19,732 testing images of the remaining 750 identities. \\
\textbf{DukeMTMC-reID}~\cite{zheng2017unlabeled} was collected in the winter of Duke University from 8 different cameras, which contains 16,522 images of 702 identities for training, and the remaining images of 702 identities for testing, including 2,228 images as query and 17,661 images as gallery. \\
\textbf{CUHK03}~\cite{li2014deepreid} contains 14,096 images which are collected by the Chinese University of Hong Kong with the images taken from only 2 cameras. Specifically, CUHK03 dataset is divided into 7,368 images of 767 identities as the training set and the remaining 5,328 images of 700 identities as the testing set. In this work, we conduct experiments on the labeled bounding boxes (CUHK03 (Labeled)). 

In our experiments, we verify the effectiveness of Colo-ReID and compare its performance with other methods on our Colo-Pair dataset and other ReID datasets respectivdely. To be more specific, we follow the standard evaluation protocol~\cite{zheng2015scalable} used in the Re-ID task, and adopt the standard evaluation metrics to evaluate the performance of object retrieval, including mean Average Precision (mAP) and Cumulative Matching Characteristics (CMC) at Rank-1, Rank-5 and Rank-10.


\subsection{Implementation Details}
In our experiment, ResNet-50~\cite{he2016deep} is regarded as the backbone. Following the training procedure in~\cite{chen2023colo}, we adopt common strategies such as random flipping and random cropping for data augmentation, and employ the Adam optimizer with a weight decay co-efficient of $5 \times 10^{-4}$ for parameter optimization. During training,
we adopt triplet loss to train the model for 10 iterations, where the hyperparameter $m$ for the loss function in Eq.~\ref{eq8} is set to 0.3. In addition, the batch size for training is set to 64. As for triplet selection, we randomly select 16 persons and sampled 4 images for each identity.
The initial learning rates for meta-training and meta-testing are set to $3.5 \times 10^{-5}$, and linearly increased to $3.5 \times 10^{-4}$ within 10 iterations. All the experiments are performed on PyTorch framework with one Nvidia GeForce RTX 2080Ti GPU on a server equipped with a Intel Xeon Gold 6130T CPU.


\subsection{Ablation study}
In order to prove the effectiveness of individual technical contributions, we perform the following ablation studies in terms of the proposed Colo-ReID method and meta-learning normalization strategy.

\textbf{The effectiveness of the Colo-ReID method.}
To validate the effectiveness of different components, we conduct several ablation experiments. It is worth noting that the ablation scheme baseline method only adopts a bare-bone ResNet-50~\cite{he2016deep}. According to Fig.~\ref{tab1}, it can be observed that the mAP accuracy drops significantly from \textbf{72.0\%} to \textbf{63.2\%}  and from \textbf{79.4\%} to \textbf{69.1\%} on Colo-Pair dataset in terms of Median performance and Maximum performance (mAP accuracy)
when the meta-learning method is not adopted. Additionally, similar drops can also be observed for other settings, \textit{e.g.} Median performance and Minimum performance. The effectiveness of the meta-learning module can be largely attributed to the fact that it enhances the model's representation capability to learn more general knowledge that can be applied to a wide range of tasks, which is vital for domain generalization when deploying deep models in real-world scenarios, especially for the medical area. 
Mathematically, the p-value is calculated using integral calculus from the area under the probability distribution curve for all values of statistics that are at least as far from the reference value as the observed value is, relative to the total area under the probability distribution curve.
More importantly, it is worth noting that all the \textit{p-values} are elaborated in the captions of Fig.~\ref{tab1}, which is adopted to measure the significance of difference between sample data and previously hypothesis, and it also indicates the likelihood of both distributions to be the same (p=1) or to be different from each other (p$\rightarrow$0).  (All p-values $<$ 0.05 (mAP: p$^{12}$ = 6.86 × 10$^{-3}$, p$^{23}$ = 2.82 × 10$^{-5}$, p$^{13}$ = 6.45 × 10$^{-7}$; Rank-1: p$^{12}$ = 1.85 × 10$^{-2}$, p$^{23}$ = 6.55 × 10$^{-5}$, p$^{13}$ = 1.22 × 10$^{-4}$; Rank-5: p$^{12}$ = 3.20 × 10$^{-2}$, p$^{23}$ = 3.46 × 10$^{-4}$, p$^{13}$ = 1.27 × 10$^{-4}$; Rank-10: p$^{12}$ = 2.14 × 10$^{-2}$, p$^{23}$ = 6.67 × 10$^{-4}$, p$^{13}$ = 5.88 × 10$^{-5}$)).

\textbf{The effectiveness of the meta-learning normalization strategy.}
To verify the effectiveness of enhanced meta-learning regularization (\textbf{MLR}), we conducted a second set of ablation experiments (Base+Colo and Base+Colo+MLR) on top of the BL+ML baseline. The second and third rows of Fig.~\ref{tab1} show the statistical results of this set of ablation experiments. It can be seen that the use of meta-learning regularization further optimized the results of colon polyp recognition in this set of experiments. For example, the meta-learning normalization strategy can improve the performance of mAP accuracy from \textbf{72.5\%} to \textbf{74.8\%} in terms of Mean performance, \textbf{72.0\%} to \textbf{74.4\%} in terms of Median performance, which significantly verifies the effectiveness of our proposed meta-learning normalization strategy.

\begin{table}[!t]
  \centering
  \caption{Performance comparison (\%) with other image retrieval or object re-identification methods on the Colo-Pair dataset. \textbf{Bold} indicates the best and \underline{underline} the second best.}
  \setlength{\tabcolsep}{1.6mm}{
    \begin{tabular}{lccccc}
    \toprule
    \multirow{2}[4]{*}{Methods} & \multirow{2}[4]{*}{Venue} & \multicolumn{4}{c}{Polyp Re-Identification Task} \\
\cmidrule{3-6}          &       & mAP   & Rank-1 & Rank-5 & Rank-10 \\
    \midrule
    MGN~\cite{wang2018learning}   & MM 18 & 65.1  & 70.9  & 76.3  & 79.7 \\
    OSNet~\cite{zhou2019omni} & ICCV 19 & 62.6  & 65.9  & 74.1  & 79.2 \\
    BN-Neck~\cite{luo2019strong} & TMM 19 & 71.4  & 74.1  & 78.5  & 82.1 \\
    Alignedreid~\cite{luo2019alignedreid++} & PR 19 & 64.2  & 68.3  & 78.0    & 80.1 \\
    st-ReID~\cite{wang2019spatial} & AAAI 19 & 70.6  & 74.6  & 81.2  & 83.5 \\
    TransReID~\cite{he2021transreid} & ICCV 21 & 66.6  & 73.3  & 78.1  & 82.1 \\
    AGW~\cite{ye2021deep}   & TPAMI 21 & 66.3  & 73.9  & 77.8  & 81.5 \\
    MixStyle~\cite{zhou2021domain} & ICLR 21 & 57.3  & 60.8  & 69.2  & 71.7 \\
    CTL~\cite{wieczorek2021unreasonable}   & ICONIP 21 & 68.9  & 71.3  & 79.2  & 82.3 \\
    ABD-Net~\cite{chen2019abd} & ICCV 21 & \underline{72.1}  & \textbf{77.6}  & \underline{82.3}  & \underline{84.6} \\
    Cluster Contrast~\cite{dai2022cluster} & ACCV 22 & 60.3  & 68.9  & 74.8  & 77.4 \\
    Nformer~\cite{wang2022nformer} & CVPR 22 & 58.3  & 60.6  & 65.7  & 73.2 \\
    AdaSP~\cite{zhou2023adaptive} & CVPR 23 & 64.0    & 67.9  & 75.7  & 81.3 \\
    \midrule
    Colo-ReID & Ours  & \textbf{74.4}  & \underline{77.5}  & \textbf{85.8}  & \textbf{90.2} \\
    \bottomrule
    \end{tabular}}%
  \label{tab2}%
\end{table}%

\begin{table}[!t]
  \centering
  \caption{Performance comparison with some representative methods by the meta learning regularization in the colonoscopy scenario. \textbf{Bold} indicates the best and \underline{underline} the second best.}
  \setlength{\tabcolsep}{1.6mm}{
    \begin{tabular}{lccccc}
    \toprule
    \multirow{2}[4]{*}{Methods} & \multirow{2}[4]{*}{Venue} & \multicolumn{4}{c}{Polyp Re-Identification Task} \\
\cmidrule{3-6}          &       & mAP   & Rank-1 & Rank-5 & Rank-10 \\
    \midrule
    MGN+MLR~\cite{wang2018learning}   & MM 18 & 66.4  & 71.3  & 77.8  & 80.3 \\
    OSNet~\cite{zhou2019omni} & ICCV 19 & 62.6  & 65.9  & 74.1  & 79.2 \\
    TransReID+MLR~\cite{he2021transreid} & ICCV 21 & 67.2  & 74.7  & 79.3  & 83.2 \\
    AGW+MLR~\cite{ye2021deep}   & TPAMI 21 & 68.2  & 74.6  & 79.2  & 83.2 \\
    ABD-Net+MLR~\cite{chen2019abd} & ICCV 21 & \underline{73.3}  & \textbf{78.4}  & \underline{84.1}  & \underline{85.2} \\
    \midrule
    Colo-ReID & Ours  & \textbf{74.4}  & \underline{77.5}  & \textbf{85.8}  & \textbf{90.2} \\
    \bottomrule
    \end{tabular}}%
  \label{tab4}%
\end{table}%

\subsection{Comparison to the state-of-the-art models}

\textbf{Polyp Re-Identification.}
In this section, we adopt the Colo-Pair dataset to perform four-fold cross-validation to verify the effectiveness of our method,
and compare our meta-learning based colon polyp recognition model Colo-ReID with state-of-the-art methods in image retrieval with the results shown in Table~\ref{tab2}.
Specifically, we applied MGN~\cite{wang2018learning}, OSNet~\cite{zhou2019omni}, BN-Neck~\cite{luo2019strong}, Alignedreid~\cite{luo2019alignedreid++}, st-ReID~\cite{wang2019spatial}, TransReID~\cite{he2021transreid},  AGW~\cite{ye2021deep}, MixStyle~\cite{zhou2021domain}, CTL~\cite{wieczorek2021unreasonable}, ABD-Net~\cite{chen2019abd}, Cluster Contrast~\cite{dai2022cluster}, and AdaSP~\cite{zhou2023adaptive} to the Colo-Pair dataset.
According to Table~\ref{tab2}, it can be observed that our model can achieve state-of-the-art performance on the Colo-Pair dataset, both in terms of Rank-5 accuracy and mAP. More importantly, by applying our meta-learning based strategy, we can obtain a performance of \textbf{74.4\%} and \textbf{85.8\%} in terms of mAP and rank-5 accuracy, leading to a significant improvement of \textbf{+2.3\%} and \textbf{+3.5\%} when compared with the second-best method ABD-Net~\cite{chen2019abd}.
It is worth mentioning that the large difference in performance with regard to mAP on all datasets is particularly noticeable. Interestingly, despite its simplicity, our architecture achieves better performance than other deep learning based approaches. Architecturally, our model is closely related to previous works such as ABD-Net~\cite{chen2019abd} and AdaSP~\cite{zhou2023adaptive}, and, in particular, st-ReID~\cite{wang2019spatial}. 

On the other hand, we also make a performance comparison between our method and other meta-learning based colon polyp recognition models for the meta-learning regularization in Table~\ref{tab4}. e.g. MGN~\cite{wang2018learning}, TransReID~\cite{he2021transreid}, AGW~\cite{ye2021deep} and ABD-Net~\cite{chen2019abd}. According to the following Table~\ref{tab4}, it can be easily observed that our Colo-ReID method can still achieve competitive performance when compared to some representative approaches in the field of image retrieval and object re-identification tasks. For example, our method outperforms the second best model ABD-Net+MLR~\cite{chen2019abd} by \textbf{+1.1\%} (74.4 vs. 73.3) and \textbf{+1.7\%} (85.8 vs. 84.1) in terms of mAP and Rank-5 accuracy respectively, which demonstrates the effectiveness and priority of our proposed meta-learning based mechanism in general medical image retrieval scenarios.

\textbf{Person Re-Identification.}
To further validate the effectiveness of our method, we also compare Colo-ReID with existing methods of CNN- or Transformer-based architecture on the re-identification task in Table~\ref{tab3}. It is worth mentioning that we do not apply any post-processing method like Re-Rank~\cite{zhong2017re} in our approach. As shown above, Colo-ReID obtains competitive performance on Market-1501, DukeMTMC-reID and CUHK03 dataset with considerable advantages when compared with previous datasets. For example, our Colo-ReID method can achieve a mAP/Rank-1 performance of 84.6\% and 87.0\% respectively on CUHK03 dataset, leading to a \textbf{+3.6\%} and \textbf{+2.3\%} improvement of mAP and Rank-1 accuracy, respectively, on the CUHK03 dataset when compared to the second-best method SCSN~\cite{chen2020salience}.

\begin{table}[!t]
  \centering
  \caption{Performance comparison (\%) with other image retrieval or object re-identification methods on the Market-1501, DukeMTMC-reID and CUHK03 dataset, respectively. \textbf{Bold} indicates the best and \underline{underline} the second best.}
  \setlength{\tabcolsep}{1.82mm}{
    \begin{tabular}{lcccccc}
    \toprule
    \multirow{2}[4]{*}{Method} & \multicolumn{2}{c}{Market-1501} & \multicolumn{2}{c}{DukeMTMC-reID} & \multicolumn{2}{c}{CUHK03} \\
\cmidrule{2-7}          & mAP   & Rank-1 & mAP   & Rank-1 & mAP   & Rank-1 \\
    \midrule
    PCB~\cite{sun2018beyond} & 81.6  & 93.8  & 69.2  & 83.3  & 57.5  & 63.7 \\
    MHN~\cite{chen2019mixed}   & 85.0  & \underline{95.1}  & 77.2  & 89.1  & 76.5  & 71.7 \\
    Pytramid~\cite{zheng2019pyramidal} & 88.2  & 95.7  & 79.0  & 89.0  & 74.8  & 78.9 \\
    BAT-net~\cite{fang2019bilinear} & 85.5  & 94.1  & 77.3  & 87.7  & 73.2  & 76.2 \\
    ISP~\cite{zhu2020identity}   & \textbf{88.6}  & 95.3  & \textbf{80.0}  & \underline{89.6}  & 71.4  & 75.2 \\
    CBDB-Net~\cite{tan2021incomplete} & 85.0  & 94.4  & 74.3  & 87.7  & 72.8  & 75.4 \\
    C2F~\cite{zhang2021coarse}   & 87.7  & 94.8  & 74.9  & 87.4  & 84.1  & 81.3 \\
    MGN~\cite{wang2018learning}   & 86.9  & \textbf{95.7}  & 78.4  & 88.7  & 66.0  & 66.8 \\
    SCSN~\cite{chen2020salience}  & \underline{88.3}  & 92.4  & \underline{79.0}  & \textbf{91.0}  & \underline{81.0}  & \underline{84.7} \\
    \midrule
    Colo-ReID & 82.1  & 93.3  & 72.3  & 85.9  & \textbf{84.6}  & \textbf{87.0}  \\
    \bottomrule
    \end{tabular}}%
  \label{tab3}%
\end{table}%

\subsection{Visualization results}
\begin{figure}[!t]
\centering{\includegraphics[width=\linewidth]{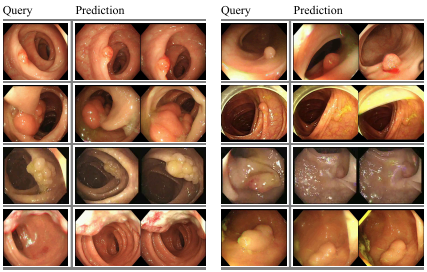}}
\caption{Qualitative visualization of ranking results of our proposed approach Colo-ReID with dynamic meta-learning regulation mechanism on Colo-ReID dataset. We obviously observe that our method shows a great robustness regardness of the viewpoint or illumination variation (self-discrepancy) of these captured polyps.}
\label{fig5}
\end{figure}

\begin{figure}[!t]
\centering{\includegraphics[width=\linewidth]{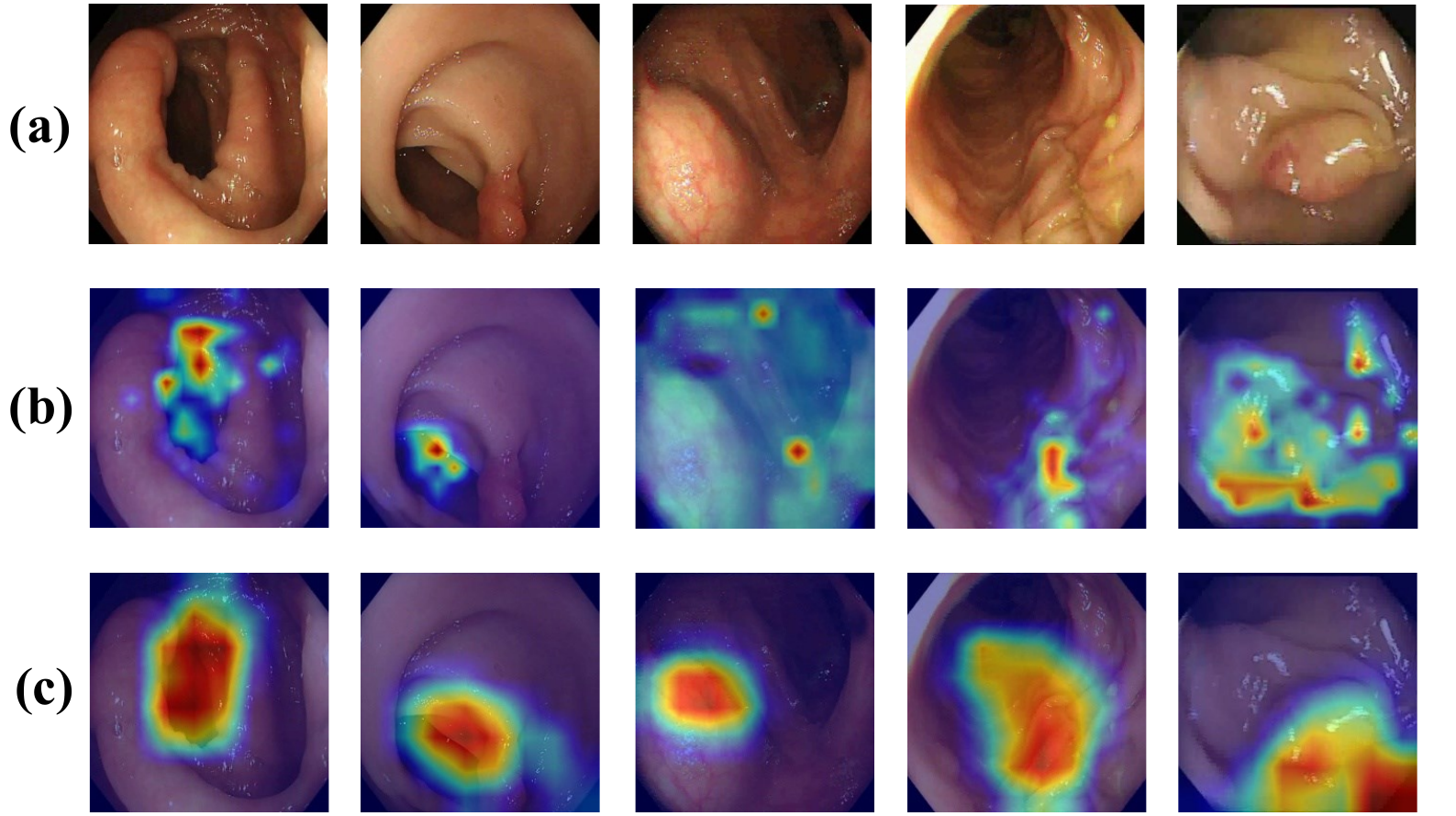}}
\caption{Visualization of attention maps with EigenGradCAM: (a) Original images; (b) Colo-ReID method without MLR regulation; (c) Our proposed Colo-ReID method on Colo-Pair dataset. It can be easily observed that enhanced meta-learning regularization method can capture global context information and more discriminative parts, which are
further enhanced in our Colo-ReID method for better performance.}
\label{fig6}
\end{figure}

To further validate the effectiveness of our meta-learning-based method, we present some visual exemplars qualitatively with respect to different real target datasets. As shown in Fig.~\ref{fig5}, it can be observed that Colo-ReID shows great robustness regardless of the viewpoint or illumination variation (referred to as ``self-discrepancy among intra-class or inter-class relations of colonoscopic polyp dataset" in this paper) of these captured polyp datasets. This phenomenon can be largely contributed to the fact that our method can help the baseline model to learn more representative, robust and discriminative visual features with better semantic understanding, which can significantly make our meta-learning based method more robust to perturbations and disturbances.
In addition, we perform in-depth analysis of feature response in Colo-ReID method, and also show some qualitative examples of EigenGradCAM~\cite{muhammad2020eigen} visualizations in Fig.~\ref{fig6}, compared to the Colo-ReID method without MLR regulation, we observe that our Colo-ReID method  attends to relevant image regions or discriminative parts for
making decisions, indicating that our Colo-ReID model can effectively explore
more global context information and meaningful visual features with better
semantic understanding, which significantly make our model more robust to
perturbations, such as light-colored areas, disturbance due to its contextual
discriminability in visual representation.

\subsection{Discussion}
According to the experiment results, the proposed Colo-ReID method has shown its potential for polyp ReID task in small sample scenarios for polyp re-identification. In continuation, an explanation is provided for two interesting phenomena observed during the experiments.

Firstly, according to Table~\ref{tab2}, there exists an interesting phenomenon that the performance of Colo-ReID is slightly inferior to the performance of ABD-Net~\cite{chen2019abd} for polyp re-identification (e.g. \textcolor[rgb]{1.00,0.39,0.09}{\textbf{77.6\%}} vs. \textcolor[rgb]{0.20,0.40,0.80}{\textbf{77.5\%}} Rank-1 on Colo-Pair dataset). This may be due to the orthogonality constraint in ABD-Net~\cite{chen2019abd} which is able to enforce the diversity on both hidden activations and weights and thus significantly improves the performance for object ReID systems.

Secondly, for the small-sample problem, the evaluation results produced under different splits of the training and testing sets can vary greatly due to the randomness of the pre-setting Colo-ReID dataset. This phenomenon is mainly due to the limited size (only 60 videos from 30 patients) of existing polyp ReID datasets since deep learning models require extensive data to learn prior knowledge from training data. Although we adopt meta-learning training strategies to improve the model's ability in some degree, the capability is still relatively limited when compared to other identity-guided methods (\textit{e.g.} ISP~\cite{zhu2020identity}) and salience-guided based methods (\textit{e.g.} SCSN~\cite{chen2020salience}). 
Consequently, these challenges warrant further research and consideration when deploying ReID models in real scenarios.

\section{Conclusion and Future Work}
\label{sec5}
In this work, we propose a simple but effective training method named Colo-ReID. Based on it, a dynamic \textbf{M}eta-\textbf{L}earning \textbf{R}egulation mechanism called MLR is introduced to further boost the performance of colonoscopic polyp ReID, which allows it to be more robust in the general scenarios.
Comprehensive experiments conducted on the Colo-Pair dataset and other related datasets demonstrate that Colo-ReID consistently improves over the baseline method by a large margin, proving itself as a strong baseline for the colonoscopic polyp ReID task. 
In the future, we will focus on learning discriminative global features and local representations with self-attention  for multi-domain colonoscopic polyp retrieval in medical community.

\section*{Acknowledgments}
This work was partially supported by the National Natural Science Foundation of China under Grant No.62301315, Startup Fund for Young Faculty at SJTU (SFYF at SJTU) under Grant No.23X010501967, Shanghai Municipal Health Commission Health Industry Clinical Research Special Project under Grant No.202340010 and 2025 Key Research Initiatives of Yunnan Erhai Lake National Ecosystem Field Observation Station under Grant No.2025ZD03.
The authors would like to thank the anonymous reviewers for their valuable suggestions and constructive criticisms.

\bibliographystyle{IEEEtran}


\bibliography{TCSVT}

\end{document}